\documentclass[10pt,twocolumn]{article}

\usepackage[T1]{fontenc}
\usepackage[utf8]{inputenc}
\usepackage{times}
\usepackage[safe]{tipa}
\usepackage[letterpaper,margin=0.75in,columnsep=0.25in]{geometry}
\usepackage{booktabs}
\usepackage{multirow}
\usepackage{amsmath}
\usepackage[numbers,sort&compress]{natbib}
\usepackage[hidelinks]{hyperref}
\usepackage{url}
\usepackage{xcolor}

\newcommand{\gm}{\textsc{Gloss-MT}}
\newcommand{\term}[1]{\texttt{<}#1\texttt{>}}

\title{\gm: Controllable Low-Resource Machine Translation\\with an Editable Dictionary}

\author{%
Mich-Seth Owusu \quad Stephen E. Moore \quad Akwasi Asare \quad Lawrence Adu Gyamfi\\
Paul Azunre \quad Joel Budu \quad Jonathan Asiamah}

\date{}

\begin{document}
\maketitle

\begin{abstract}
Adapting a pretrained machine translation (MT) model to new terminology usually requires new parallel data and another round of fine-tuning, which is costly for low-resource languages. We ask whether controllability can be added to MT, and the fine-tuning burden reduced, without degrading the pretrained model. \gm{} looks up the source's noun phrases in an offline English--Twi dictionary, replaces every phrase found by its Twi translation in place, and fine-tunes a translation model once to translate these mixed inputs. We apply the same data and pipeline to NLLB-200 (600M, 1.3B, 3.3B), OPUS-MT en-tw and TranslateGemma 4B. On a document-level English$\rightarrow$Asante Twi benchmark (84 documents, four domains), every fine-tune scores above its pretrained base in chrF2 (+2.85 to +5.48), and the three NLLB fine-tunes are the best open systems (50.12--50.80). A direct controllability test on human-translated Bible passages shows that the three fine-tunes tested write 79--97\% of the dictionary terms, and that replacing every dictionary entry by a different term changes the output accordingly (70--97\% of the new terms appear) with no retraining. We release the models, data and code.
\end{abstract}

\section{Introduction}

Twi (Akan) has tens of millions of speakers but little parallel text. Pretrained multilingual models such as NLLB-200~\cite{nllb2022} cover it, and fine-tuning them on a few thousand in-domain pairs is known to help~\cite{adelani2022few}. Yet despite strong benchmark performance, such models have seen little production use for low-resource languages. A recurring reason is vocabulary: they often lack the correct words for the everyday entities of the settings they serve, such as local institutions, government programmes, products and places. These languages are living, predominantly spoken languages, and new names for entities enter use continuously; fine-tuning cannot keep up, since each new term or domain convention (a ministry's budget terms, legal vocabulary, a dialectal preference) needs new parallel data and new training. Bilingual wordlists and glossaries, in contrast, are the resource that language communities and field linguists most often produce, and they can be updated as quickly as the language changes.

We study the question: \emph{can we add controllability to machine translation, and reduce the burden of fine-tuning, without degrading the performance of pretrained MT models?} It has three parts: (i)~\textbf{control}: can an operator change which term a phrase receives by editing an external dictionary instead of retraining? (ii)~\textbf{a light fine-tune}: can one short fine-tune teach a model to use dictionary terms, so that later changes need only dictionary entries? (iii)~\textbf{no degradation}: does the model translate at least as well as the pretrained model it starts from?

\gm{} answers with a simple pipeline: dictionary terms are pretranslated \emph{in place} in the source, and the model is fine-tuned once to keep them. Our contributions are: (1)~the pipeline and a training set built by aligning English noun phrases to their spans in gold Twi; (2)~fine-tunes of five pretrained models of three families, each compared with its own base on an independent document-level benchmark; (3)~a direct controllability test with a counterfactual (``swap'') dictionary, on human-translated references; and (4)~released models, data and code.\footnote{Models: \url{https://huggingface.co/ghanaopenai}; code: \url{https://github.com/GhanaOpenAI/gloss-mt}; benchmark: \url{https://github.com/GhanaNLP/nsanku-mt-benchmark}.}

\section{Related Work}

Terminology can be imposed at decoding time with lexically constrained beam search~\cite{hokamp2017lexically,post2018fast}, which guarantees the terms but slows decoding and handles inflection poorly. Alternatively, the model is trained to use terms annotated in its input: \citet{dinu2019training} append target terms inline next to the source words, \citet{song2019code} replace source phrases by their pre-specified translations (code-switching), and \citet{bergmanis2021facilitating} annotate target lemmas so the model can inflect them. \citet{ghazvininejad2023dictionary} prompt LLMs with dictionary entries for rare words. \gm{} follows the in-place replacement idea of \citet{song2019code}, applies it with a large community-style dictionary rather than a few constraints per sentence, and asks a different question: whether a \emph{single} fine-tune of an existing pretrained MT model makes the dictionary an operator's control surface without costing translation quality. For Twi, prior work built English--Twi parallel data~\cite{azunre2021english} and fine-tuned pretrained models on news translation~\cite{adelani2022few}.

\section{Method}

\paragraph{Dictionary.} An offline English$\rightarrow$Twi dictionary of 613,259 phrases with exactly one translation each. Entries come from machine-translated phrase lists (re-translated with Gemini where an entry had several candidates) and from the phrase--span alignments of the training data below.

\paragraph{Pipeline.} A document is split into lines, and consecutive lines are packed into chunks of at most 192 source tokens (128 for OPUS-MT), cutting only at line boundaries so that Markdown structure is preserved. The chunk's noun phrases are extracted with spaCy~\cite{honnibal2020spacy} and looked up; each phrase found is replaced in the English, longest first, by its Twi translation in angle brackets, and phrases not found stay English. For example, ``\emph{Our people have courage, but courage without foundation is foolishness.}'' becomes ``\emph{Our \term{nnipa} have \term{akokoduro}, but \term{akokoduro} without \term{fapem} is \term{nkwasias{\textepsilon}m}.}'' The model writes the terms in brackets too, and the pipeline strips them. Table-of-contents dot leaders are kept out of the model and restored.

\paragraph{Training data.} From 15,134 English--Twi training documents we build 31,909 line-aligned chunks. Gemini aligns each English noun phrase of a chunk to its span in the reference Twi; a span is kept only if it occurs verbatim on the phrase's line. Each aligned pair is bracketed on both sides: the input carries the Twi span in place of the phrase, and the target carries the same span in brackets. The model thus learns to copy the bracketed terms and translate the sentence around them. The data is released as \texttt{gloss-mt-parallel-en-twi}.

\paragraph{Models.} We fine-tune NLLB-200 3.3B with LoRA~\cite{hu2022lora} (r=32 on all attention and feed-forward projections, 2\% of the parameters, merged after training); NLLB-200 1.3B and distilled 600M with full fine-tuning (Adafactor~\cite{shazeer2018adafactor}, lr $10^{-4}$); OPUS-MT en-tw~\cite{tiedemann2020opus} ($\sim$75M parameters, full fine-tuning, AdamW, lr $10^{-4}$); and TranslateGemma 4B,\footnote{\url{https://huggingface.co/google/translategemma-4b-it}} a translation-specialised Gemma 3 model, with LoRA (r=32), using its own prompt with target code \texttt{ak} since it has no Twi code. A literal new-line token is added to the NLLB tokenizer. Every 1,000 steps the model translates 200 fixed validation chunks, and the checkpoint with the best mean of BLEU, chrF2 and $100\,e^{-\mathrm{loss}}$ is kept, with early stopping. NLLB and OPUS-MT decode with beam 4, TranslateGemma greedily.

\section{Evaluation}

\paragraph{Benchmark.} The nsanku MT benchmark scores document-level chrF2~\cite{popovic2015chrf} and BLEU~\cite{papineni2002bleu} (sacreBLEU~\cite{post2018call}) over 84 English$\rightarrow$Asante Twi documents in four datasets: three Citizens' Budget documents (Finance, 2021--2023), the Universal Declaration of Human Rights (UDHR), five Watchtower study editions (2026), and 75 Akan folktales (Rattray, 1930). Because 75 documents are folktales, the language score is the mean of the four per-dataset scores. Each fine-tune is compared with its pretrained base, translated with the same chunking; commercial systems are references.

\paragraph{Controllability.} Each test set is translated three times by the same model, changing only the dictionary: \emph{dict} (as deployed), \emph{none} (no dictionary) and \emph{swap}, a counterfactual dictionary in which every phrase found in the set receives a different real Twi term from the set's other terms, chosen so that it occurs neither in the source nor in the dict or none outputs of that passage; a swapped term can therefore only reach the output through the dictionary. A term counts as used when its words appear in order as whole words of the output. We report \textbf{adherence} (share of inserted terms in the output), \textbf{natural rate} (the same terms without the dictionary), \textbf{swap adherence} (share of swapped-in terms in the swap output) and \textbf{leak} (share of original terms surviving the swap). The main test set is 300 passages of three consecutive Bible verses, with the Good News Translation as English and the human Asante Twi \emph{Nkwa As\textepsilon m} as reference (2,766 inserted terms); no passage shares an 8-word span with the training data. Both versions are free, meaning-based translations, so absolute scores are low and the with/without comparison is what matters. We also run the test on 300 validation chunks (5,387 terms) whose references are not all human-curated.

\begin{table*}[t]
\centering
\small
\setlength{\tabcolsep}{5pt}
\begin{tabular}{llcccccc}
\toprule
& System & chrF2 & BLEU & Finance & UDHR & Watchtower & Folktales \\
\midrule
\multirow{2}{*}{References}
& Gemini 3.8 Flash & 54.64 & 17.25 & 52.50 & 56.44 & 69.28 & 40.33 \\
& Google Translate & 51.26 & 15.67 & 48.06 & 52.30 & 67.15 & 37.54 \\
\midrule
\multirow{2}{*}{NLLB-200 3.3B} & base & 47.74 & \textbf{14.15} & 45.25 & \textbf{52.05} & 56.89 & 36.76 \\
& \gm{} & \textbf{50.80} \small(+3.06) & 13.11 & \textbf{50.23} & 51.66 & \textbf{64.17} & \textbf{37.16} \\
\midrule
\multirow{2}{*}{NLLB-200 1.3B} & base & 47.42 & \textbf{13.68} & 45.72 & 50.47 & 57.46 & 36.05 \\
& \gm{} & \textbf{50.27} \small(+2.85) & 12.82 & \textbf{49.41} & \textbf{51.51} & \textbf{63.66} & \textbf{36.51} \\
\midrule
\multirow{2}{*}{NLLB-200 600M} & base & 46.71 & \textbf{12.74} & 44.10 & 50.17 & 57.06 & 35.50 \\
& \gm{} & \textbf{50.12} \small(+3.41) & 12.23 & \textbf{49.81} & \textbf{51.25} & \textbf{62.80} & \textbf{36.61} \\
\midrule
\multirow{2}{*}{OPUS-MT en-tw} & base & 42.52 & 9.71 & 37.78 & 47.52 & 54.23 & 30.53 \\
& \gm{} & \textbf{48.00} \small(+5.48) & \textbf{10.91} & \textbf{46.13} & \textbf{50.09} & \textbf{60.62} & \textbf{35.15} \\
\midrule
TranslateGemma 4B & \gm{} & 49.81 & 12.24 & 50.03 & 51.19 & 62.54 & 35.48 \\
\bottomrule
\end{tabular}
\caption{English$\rightarrow$Asante Twi on the nsanku benchmark (84 documents; chrF2 per dataset; the mean is over the four datasets). Bold: better of base and \gm{}. Zero-shot, TranslateGemma does not produce Twi, so it has no base row. MADLAD-400 3B~\cite{kudugunta2023madlad}, the remaining open baseline, scores 37.14.}
\label{tab:bench}
\end{table*}

\section{Results}

\paragraph{No degradation in chrF2.} Table~\ref{tab:bench} shows that every fine-tune scores above its pretrained base in chrF2, by +2.85 to +5.48. The gain concentrates in contemporary prose: Watchtower rises by 5.7--7.3 points for the NLLB models and Finance by 3.7--5.7. The only domain below its base is the UDHR for NLLB 3.3B ($-0.39$). The 1930 folktales, written in pre-standard orthography, gain least. The three NLLB fine-tunes are the best open systems, behind only the commercial Gemini 3.8 Flash and Google Translate. OPUS-MT gains most (+5.48) and, at 75M parameters, scores above the base NLLB 3.3B (48.00 vs 47.74).

\paragraph{Model size matters little once fine-tuned.} The fine-tuned NLLB models span only 50.12--50.80 from 600M to 3.3B, and the 600M fine-tune scores 2.38 above the base 3.3B model, so the cheapest model is a practical deployment choice.

\paragraph{BLEU.} The NLLB fine-tunes are slightly below their bases in BLEU ($-0.51$ to $-1.04$), mostly on the UDHR and Watchtower, while chrF2 rises on both; OPUS-MT gains BLEU (+1.20). The fine-tuned output matches more of the reference's characters but fewer of its exact word sequences, plausibly because dictionary terms differ from the reference's word choices. We report this as a limit on the no-degradation claim.

\paragraph{Validation scores do not predict the benchmark.} TranslateGemma scored above NLLB 3.3B on validation (chrF2 64.96 vs 63.66) but below it on the benchmark, and the 600M model above the 1.3B on validation (63.99 vs 62.14) but slightly below it on the benchmark. Checkpoint selection on in-domain validation data is a weak proxy for document-level quality.

\begin{table}[t]
\centering
\footnotesize
\setlength{\tabcolsep}{3pt}
\begin{tabular}{lccccc}
\toprule
Model & Adh. & Nat. & Swap & Leak & chrF2 d/n \\
\midrule
\multicolumn{6}{l}{\emph{Asante Twi Bible (human references)}} \\
\gm{} TranslateGemma & \textbf{96.6} & 26.2 & \textbf{97.3} & 1.9 & 37.2 / 32.1 \\
\gm{} NLLB 3.3B & 94.4 & 38.4 & 95.2 & 2.2 & 41.2 / 40.9 \\
\gm{} OPUS-MT & 79.2 & 32.5 & 69.7 & 1.7 & 36.4 / 33.6 \\
NLLB 3.3B base & 66.0 & 33.9 & 72.0 & 2.8 & 38.9 / 39.6 \\
OPUS-MT base & 1.9 & 30.3 & 0.4 & 1.4 & 19.1 / 30.1 \\
\midrule
\multicolumn{6}{l}{\emph{Validation chunks}} \\
\gm{} TranslateGemma & \textbf{97.8} & 36.1 & \textbf{95.9} & 6.1 & 66.0 / 67.7 \\
\gm{} NLLB 3.3B & 88.4 & 37.8 & 75.8 & 6.6 & 65.4 / 71.0 \\
\gm{} OPUS-MT & 76.2 & 34.7 & 61.0 & 5.6 & 59.2 / 64.3 \\
NLLB 3.3B base & 55.8 & 26.9 & 49.0 & 3.6 & 41.3 / 45.5 \\
OPUS-MT base & 1.8 & 13.9 & 0.1 & 1.3 & 15.0 / 26.1 \\
\bottomrule
\end{tabular}
\caption{Controllability (\%). Adh.: inserted terms in the output; Nat.: the same terms without the dictionary; Swap: swapped-in terms in the output; Leak: original terms surviving the swap; chrF2 with / without the dictionary.}
\label{tab:ctrl}
\end{table}

\paragraph{The dictionary controls the fine-tuned models.} We test three fine-tunes, one per model family, and two of their bases. On the Bible (Table~\ref{tab:ctrl}), \gm{} TranslateGemma and NLLB 3.3B write 95--97\% of the inserted terms, against 26--38\% of the same terms without the dictionary. After every entry is replaced, they write 95--97\% of the new terms and about 2\% of the original ones survive: editing the dictionary changes the translation as intended, without retraining. TranslateGemma follows multi-word terms as reliably as single words (98.1\% vs 97.7\% on validation). OPUS-MT, the smallest model, follows 70\% of swapped terms; the swapped terms are often wrong in context, and a small model drops them more often.

\paragraph{Fine-tuning creates the control.} Base OPUS-MT ignores the bracketed terms (1.9\%) and degrades when they are present (chrF2 30.1$\rightarrow$19.1). Base NLLB 3.3B copies some Twi from its input (66.0\%), and fine-tuning raises adherence to 94.4\%.

\paragraph{No quality cost on human references.} On the Bible, translating with the dictionary scores at least as well in chrF2 as without it: TranslateGemma +5.0, OPUS-MT +2.8, NLLB 3.3B +0.3 (BLEU 10.75 vs 10.91). This holds although only 35.9\% of the inserted terms occur in the human reference. On the validation chunks the dictionary instead \emph{lowers} chrF2 for every fine-tune ($-1.7$ to $-5.6$); the fine-tuned models reproduce those references' style more closely without the dictionary, which suggests the drop reflects the references rather than a loss of quality, and is why we rely on the human-referenced set for this comparison.

\section{Limitations}

Control is learned, not enforced: adherence is high but not complete, and lower for the smallest model. Term matching is by exact whole words, so a term that Twi morphology changes (a prefix or plural) counts as missed and adherence is a lower bound; target-lemma annotation~\cite{bergmanis2021facilitating} could address inflection. The dictionary is largely machine-generated, and its entries often differ from the human references' choices. The fine-tuned NLLB models lose some BLEU. The human-referenced controllability set is a single domain (the Bible), translated freely on both sides, and we have not yet measured the quality gain from adding curated terminology for a new domain. All evaluation is English$\rightarrow$Asante Twi.

\section{Conclusion}

A single fine-tune on dictionary-pretranslated inputs makes an external dictionary an effective control surface for pretrained MT models: the three fine-tunes we tested follow 79--97\% of the dictionary terms and follow edited entries without retraining, and all five fine-tunes, from three model families, score above their pretrained bases in chrF2 on an independent document-level benchmark, with a small BLEU cost for the NLLB models. Because the gain is similar from 600M to 3.3B parameters, communities can obtain controllable translation from a small model and maintain it by editing wordlists rather than collecting parallel data. Models, data and code are released.

\bibliographystyle{plainnat}
\bibliography{refs}

\end{document}